\documentclass[11pt,a4paper]{article}
\usepackage[hyperref]{acl2017}
\usepackage{times}
\usepackage{latexsym}

\usepackage{url}

\aclfinalcopy % Uncomment this line for the final submission
\def\aclpaperid{44} %  Enter the acl Paper ID here

\usepackage{booktabs}
\usepackage{graphicx}
\usepackage{tikz}

\title{Macquarie University at BioASQ 5b -- Query-based Summarisation
  Techniques for Selecting the Ideal Answers}

\author{Diego Moll\'{a}\\
  Department of Computing\\
  Macquarie University\\
  Sydney, Australia\\
  {\tt diego.molla-aliod@mq.edu.au} \\}

\date{}

\begin{document}
\maketitle
\begin{abstract}
  Macquarie University's contribution to the BioASQ challenge (Task 5b
  Phase B) focused on the use of query-based extractive summarisation
  techniques for the generation of the ideal answers. Four runs were
  submitted, with approaches ranging from a trivial system that
  selected the first $n$ snippets, to the use of deep learning
  approaches under a regression framework. Our experiments and the
  ROUGE results of the five test batches of BioASQ indicate
  surprisingly good results for the trivial approach. Overall, most of
  our runs on the first three test batches achieved the best ROUGE-SU4
  results in the challenge.
\end{abstract}

\section{Introduction}

The main goal of query-focused multi-document summarisation is to
summarise a collection of documents from the point of view of a
particular query. In this paper we compare the use of various
techniques for query-focused summarisation within the context of the
BioASQ challenge. The BioASQ challenge \cite{Tsatsaronis:2015} started
in 2013 and it comprises various tasks centred on biomedical semantic
indexing and question answering. The fifth run of the BioASQ challenge
\cite{Nentidis:2017}, in particular, had three tasks:

\begin{itemize}
\item BioASQ 5a: Large-scale online biomedical semantic indexing.
\item BioASQ 5b: Biomedical semantic question answering. This task had
  two phases:
  \begin{itemize}
  \item Phase A: Identification of relevant information.
  \item Phase B: Question answering. 
  \end{itemize}
\item BioASQ 5c: Funding information extraction from biomedical literature.
\end{itemize}

The questions used in BioASQ 5b were of three types: yes/no, factoid,
list, and summary. Submissions to the challenge needed to provide an
exact answer and an ideal answer. Figure~\ref{fig:data} shows examples of
exact and ideal answers for each type of question.
\begin{figure*}
  \begin{description}
  \item[yes/no]   Does Apolipoprotein E (ApoE) have anti-inflammatory activity?
  \begin{itemize}
  \item \textbf{Exact answer}: yes
  \item \textbf{Ideal answer}: Yes. ApoE has anti-inflammatory activity
  \end{itemize}
\item[factoid] Which type of lung cancer is afatinib used for?
  \begin{itemize}
  \item \textbf{Exact answer}: EGFR-mutant non small cell lung carcinoma
  \item \textbf{Ideal answer}: Afatinib is a small molecule covalently binding
    and inhibiting the EGFR, HER2 and HER4 receptor tyrosine
    kinases. Trials showed promising efficacy in patients with
    EGFR-mutant NSCLC or enriched for clinical benefit from EGFR
    tyrosine kinase inhibitors gefitinib or erlotinib. 
  \end{itemize}
\item[list] Which are the Yamanaka factors?
  \begin{itemize}
  \item \textbf{Exact answer}: [OCT4, SOX2, MYC, KLF4]
  \item \textbf{Ideal answer}: The Yamanaka factors are the OCT4, SOX2, MYC, and KLF4 transcription factors
  \end{itemize}
\item[summary] What is the role of brain natriuretic peptide in
  traumatic brain injury patients ?
  \begin{itemize}
  \item \textbf{Exact answer}: N/A
  \item \textbf{Ideal answer}: Brain natriuretic peptide concentrations are
    elevated in patients with traumatic brain during the acute phase
    and correlate with poor outcomes. In traumatic brain injury
    patients higher brain natriuretic peptide concentrations are
    associated with more extensive SAH,  elevated ICP and
    hyponatremia. Brain natriuretic peptide may play an adaptive role
    in recovery through augmentation of cerebral blood flow. 
  \end{itemize}
  \end{description}
  
  \caption{Examples of questions with their exact and ideal answers in
  BioASQ 5b.}
  \label{fig:data}
\end{figure*}
We can see that the ideal answers are full sentences that expand the
information provided by the exact answers. These ideal answers could
be seen as the result of query-focused multi-document
summarisation. We therefore focused on Task 5b Phase B, and in that
phase we did not attempt to provide exact answers. Instead, our runs
provided the ideal answers only.

% We ran multiple experiments with various approaches to query-focused
% multi-document summarisation. 
In this paper we will describe the
techniques and experiment results that were most relevant to our final
system runs. Some of our runs were very simple, yet our preliminary
experiments revealed that they were very effective and, as expected,
the simpler approaches were much faster than the more complex
approaches.

Each of the questions in the BioASQ test sets contained the text of
the question, the question type, a list of source documents, and a
list of relevant snippets from the source documents. We used this
information, plus the source documents which are PubMed abstracts
accessible using the URL provided in the test sets.

Overall, the summarisation process of our runs consisted of the
following two steps:

\begin{enumerate}
\item Split the input text (source documents or snippets) into
  candidate sentences and score each candidate sentence.
\item Return the $n$ sentences with highest score.
\end{enumerate}

The value of $n$ was determined empirically and it depended on the
question type, as shown in Table~\ref{tab:n}.

\begin{table}
  \centering
  \begin{tabular}{ccccc}
   
    & \textbf{Summary} & \textbf{Factoid} & \textbf{Yesno} & \textbf{List}\\
    \midrule
    \textbf{n} & 6 & 2 & 2 & 3
  \end{tabular}
  \caption{Value of $n$ (the number of sentences returned as the
    ideal answer) for each question type.}\label{tab:n}
\end{table}

\section{Simple Runs}\label{sec:simple}

As a first baseline, we submitted a run labelled \textbf{trivial} that
simply returned the first $n$ snippets of each question. The reason
for this choice was that, in some of our initial experiments, we
incorporated the position of the snippet as a feature for a machine
learning system. In those experiments, the resulting system did not
learn anything and simply returned the input snippets
verbatim. Subsequent experiments revealed that a trivial baseline that
returned the first snippets of the question was very hard to beat. In
fact, for the task of summarisation of other domains such as news, it
has been observed that a baseline that returns the first sentences
often outperformed other methods \cite{Brandow1995}.

As a second baseline, we submitted a run labelled \textbf{simple} that
selected the $n$ snippets what were most similar to the question. We
used cosine similarity, and we tried two alternatives for computing
the question and snippet vectors:

\begin{description}
\item[tfidf-svd:] First, generate the $tf.idf$ vector of the question
  and the snippets. We followed the usual procedure, and the $tf.idf$
  vectors of these sentences are bag-of-word vectors where each
  dimension represents the $tf.idf$ of a word. Then, reduce the
  dimensionality of the vectors by selecting the first 200 components
  after applying Singular Value Decomposition. In contrast with a
  traditional approach to generate the $tf.idf$ (and SVD) vectors
  where the statistics are based on the input text solely (question
  and snippets in our case), we used the text of the question and the
  text of the ideal answers of the training data.\footnote{In
    particular, we used the ``TfidfVectorizer'' module of the sklearn
    toolkit (\url{http://scikit-learn.org}) and fitted it with the
    list of questions and ideal answers. We then used the
    ``TruncatedSVD'' module and fitted it with the tf.idf vectors of
    the list of questions and ideal answers.} The reason for using
  this variant was based on empirical results during our preliminary
  experiments.
\item[word2vec:] Train Word2Vec \cite{Mikolov:2013} using a set of over 10
  million PubMed abstracts provided by the organisers of BioASQ. Using
  these pre-trained word embeddings, look up the word embeddings of
  each word in the question and the snippet. The vector representing a
  question (or snippet) is the sum of embeddings of each word in the
  question (or snippet). The dimension of the word embeddings was set
  to 200.
\end{description}

Table~\ref{tab:simple} shows the F1 values of ROUGE-SU4 of the
resulting summaries. The table shows the mean and the standard
deviation of the evaluation results after splitting the training data
set for BioASQ 5b into 10 folds (for comparison with the approaches
presented in the following sections).
\begin{table}
  \centering
  \begin{tabular}{lccc}
    &\textbf{trivial}&\multicolumn{2}{c}{\textbf{simple}}\\
    &                &\textbf{tfidf-svd}&\textbf{word2vec}\\
\midrule
\textbf{Mean F1}&0.2157&0.1643&0.1715\\
\textbf{Stdev F1}&0.0209&0.0097&0.0128
  \end{tabular}
  \caption{ROUGE-SU4 of the simple runs.}
  \label{tab:simple}
\end{table}

We observe that the trivial run has the best
results, and that the run that uses word2vec is second best. Our run
labelled ``simple'' therefore used cosine similarity of the sum of
word embeddings returned by word2vec.

\section{Regression Approaches}\label{sec:regression}

For our run labelled \textbf{regression}, we experimented with the use
of Support Vector Regression (SVR). The regression setup and features
are based on the work by \newcite{Malakasiotis2015}, who reported the
best results in BioASQ 3b (2015).

The target scores used to train the SVR system were the F1 ROUGE-SU4
score of each individual candidate sentence.

In contrast with the simple approaches described in
Section~\ref{sec:simple}, which used the snippets as the input data,
this time we used all the sentences of the source abstracts. We also
incorporated information about whether the sentence was in fact a
snippet as described below.

As features, we used:

\begin{itemize}
\item $tf.idf$ vector of the candidate sentence. In contrast with the
  approach described in Section~\ref{sec:simple}, The statistics used
  to determine the $tf.idf$ vectors were based on the text of the
  question, the text of the ideal answers, and the text of the
  snippets.
\item Cosine similarity between the $tf.idf$ vector of the
  question and the $tf.idf$ vector of the candidate sentence.
\item The smallest cosine similarity between the $tf.idf$ vector of
  candidate sentence and the $tf.idf$ vector of each of the snippets
  related to the question. Note that this feature was not used by
  \newcite{Malakasiotis2015}.
\item Cosine similarity between the sum of word2vec embeddings of the
  words in the question and the word2vec embeddings of the words in
  the candidate sentence. As in our run labelled ``simple'', we used
  vectors of dimension 200.
\item Pairwise cosine similarities between the words of the question
  and the words of the candidate sentence. As in the work by
  \newcite{Malakasiotis2015}, we used word2vec to compute the word
  vectors. These word vectors were the same as used in
  Section~\ref{sec:simple}. We then computed the pairwise cosine
  similarities and selected the following features:
  \begin{itemize}
  \item The mean, median, maximum, and minimum of all pairwise cosine similarities.
  \item The mean of the 2 highest, mean of the 3 highest, mean of the
    2 lowest, and mean of the 3 lowest.
  \end{itemize}
\item Weighted pairwise cosine similarities, also based in the work by
  \newcite{Malakasiotis2015}. In particular, now each word vector was
  multiplied by the $tf.idf$ of the word, we computed the pairwise
  cosine similarities, and we used the mean, median, maximum, minimum,
  mean of 2 highest, mean of 3 highest, mean of 2 lowest, and mean of
  3 lowest.
\end{itemize}

\begin{figure}
  \centering
\includegraphics[width=\columnwidth]{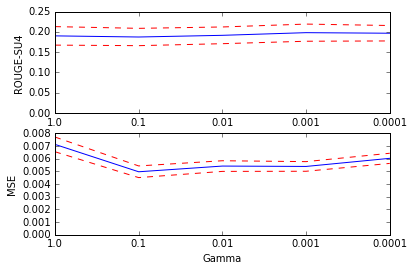}
\caption{Grid search of the Gamma parameter for the experiments using
  Support Vector Regression. The continuous lines indicate the mean of
  10-fold cross-validation over the training data set of BioASQ
  5b. The dashed lines indicate 2 $\times$ the standard deviation.}
  \label{fig:svr}
\end{figure}

Figure~\ref{fig:svr} shows the result of grid search by varying the
$gamma$ parameter of SVR, fixing $C$ to 1.0, and using the $RBF$
kernel.\footnote{We used the Scikit-learn Python package.} The figure
shows the result of an extrinsic evaluation that reports the F1
ROUGE-SU4 of the final summary, and the result of an intrinsic
evaluation that reports the Mean Square Error (MSE) between the target
and the predicted SU4 of each individual candidate sentence.

\begin{figure*}
  \centering
\includegraphics[width=\textwidth]{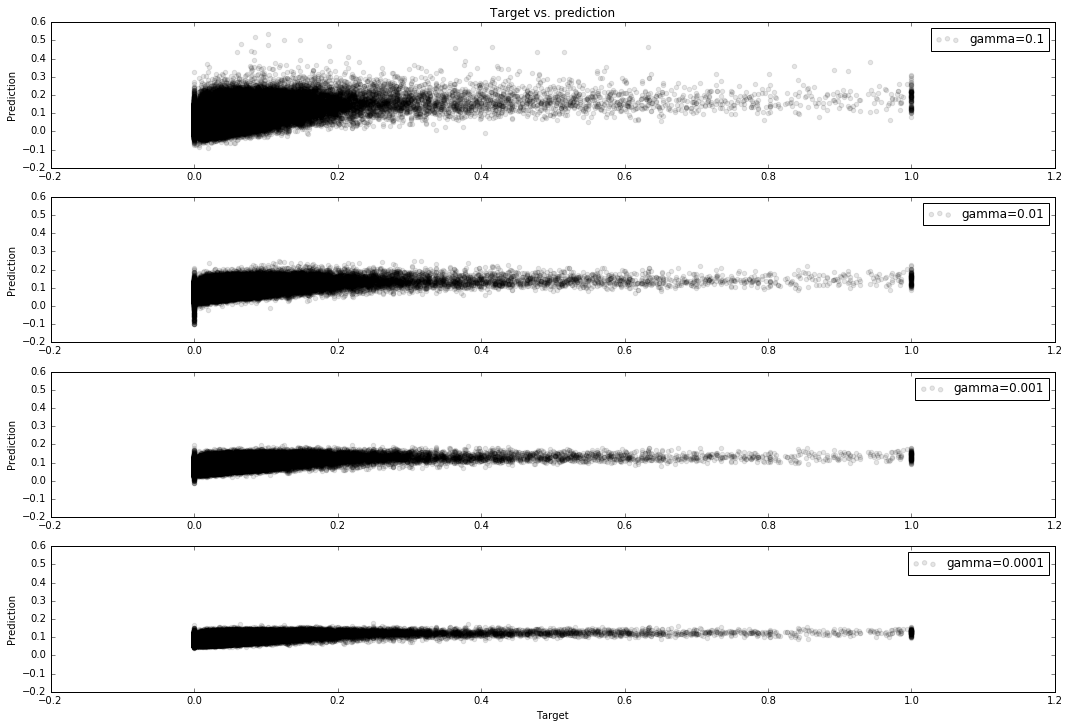}  
\caption{Target vs. predicted SU4 in the SVR experiments for various
  values of $gamma$.}
  \label{fig:scatterplot}
\end{figure*}

We can observe discrepancy between the results of the intrinsic and
the extrinsic evaluations. This discrepancy could be due to the fact
that the data are highly imbalanced in the sense that most annotated
SU4 scores in the training data have low values. Consequently, the
regressor would attempt to minimise the errors in the low values of
the training data at the expense of errors in the high values. But the
few sentences with high SU4 scores are most important for the final
summary, and these have higher prediction error. This can be observed
in the scatter plot of Figure~\ref{fig:scatterplot}, which plots the
target against the predicted SU4 in the SVR experiments for each value
of $gamma$. The SVR system has learnt to predict the low SU4 scores
to some degree, but it does not appear to have learnt to discriminate
among SU4 scores over a value of 0.4.

Our run labelled ``regression'' used $gamma=0.1$ since it gave the
best MSE in our intrinsic evaluation, and Figure~\ref{fig:scatterplot}
appeared to indicate that the system learnt best.

\section{Deep Learning Approaches}

For our run labelled \textbf{nnr} we experimented with the use of deep
learning approaches to predict the candidate sentence scores under a
regression setup. The regression setup is the same as in
Section~\ref{sec:regression}.

\begin{figure*}
  \centering
  \begin{tikzpicture}[scale=0.4]
% input
    \draw (0,0) rectangle (1,5) (0,1) -- (1,1) (0,3) -- (1,3) (0,4) -- (1,4);
    \draw (0,-7) rectangle (1,-2) (0,-6) -- (1,-6) (0,-4) -- (1,-4) (0,-3) -- (1,-3);
    \draw (-1,2.5) node[rotate=90] {sentence};
    \draw (-1,-4.5) node[rotate=90] {question};

% word embeddings
    \draw (4,-1) node [circle,draw,align=center,text width=1.7cm] (em) {embedding matrix};
    \draw (8.5,6) node {word embeddings};
    \draw (7,0) rectangle (10,5) (7,1) -- (10,1) (7,3) -- (10,3) (7,4) -- (10,4) (8,0) -- (8,5) (9,0) -- (9,5);
    \draw (7,-7) rectangle (10,-2) (7,-6) -- (10,-6) (7,-4) -- (10,-4) (7,-3) -- (10,-3) (8,-7) -- (8,-2) (9,-7) -- (9,-2);

    \draw[->] (1,2.5) -- (em);
    \draw[->] (1,-4.5) -- (em);

    \draw[->] (em) -- (7,2.5);
    \draw[->] (em) -- (7,-4.5);
% sentence embeddings
    \draw (14,2.5) node [circle,draw,align=center,text width=1.5cm] (sr) {sentence reduction};
    \draw (14,-4.5) node [circle,draw,align=center,text width=1.5cm] (qr) {question reduction};
    \draw (18,6) node {sentence embeddings};
    \draw (18,0) rectangle (19,5) (18,1) -- (19,1) (18,3) -- (19,3) (18,4) -- (19,4);
    \draw (18,-7) rectangle (19,-2) (18,-6) -- (19,-6) (18,-4) -- (19,-4) (18,-3) -- (19,-3);

    \draw[->] (10,2.5) -- (sr);
    \draw[->] (sr) -- (18,2.5);
    \draw[->] (10,-4.5) -- (qr);
    \draw[->] (qr) -- (18,-4.5);

% similarity
    \draw (21,-4.5) node [circle,draw] (t) {$\times$};
    \draw (23,-7) rectangle (24,-2) (23,-6) -- (24,-6) (23,-4) -- (24,-4) (23,-3) -- (24,-3);

    \draw[->] (19,2.5) -| (t);
    \draw[->] (19,-4.5) -- (t);

    \draw[->] (21,2.5) |- (25,1.5);
    \draw[->] (t) -- (23,-4.5);
    \draw (24,-4.5) -| (24.5,-3.5);
    \draw[->] (24.5,-3.5) -- (25,-3.5);

    \draw (21,-6.2) node {\small similarity};

% hidden layer
    \draw (25,-1) rectangle (26,4);
    \draw (25,-1) rectangle (26,-6);
    \draw (28,-1) circle[radius=1] (27.5,-1.5) -- (28,-1.5) -- (28.5,-0.5);
    \draw (28,1) node {relu};

    \draw (30,-3.5) rectangle (31,1.5);

    \draw[->] (26.2,-1) -- (27,-1);
    \draw[->] (29,-1) -- (30,-1);

% final layer
    \draw (33,-1) circle[radius=1] (32.5,-1.5) -- (33.5,-0.5);
    \draw (35,-1.5) rectangle (36,-0.5);
    \draw (33,1) node {linear};

    \draw[->] (31,-1) -- (32,-1);
    \draw[->] (34,-1) -- (35,-1);
  \end{tikzpicture}

  \caption{Architecture of the regression system.}
  \label{fig:nnr}
\end{figure*}
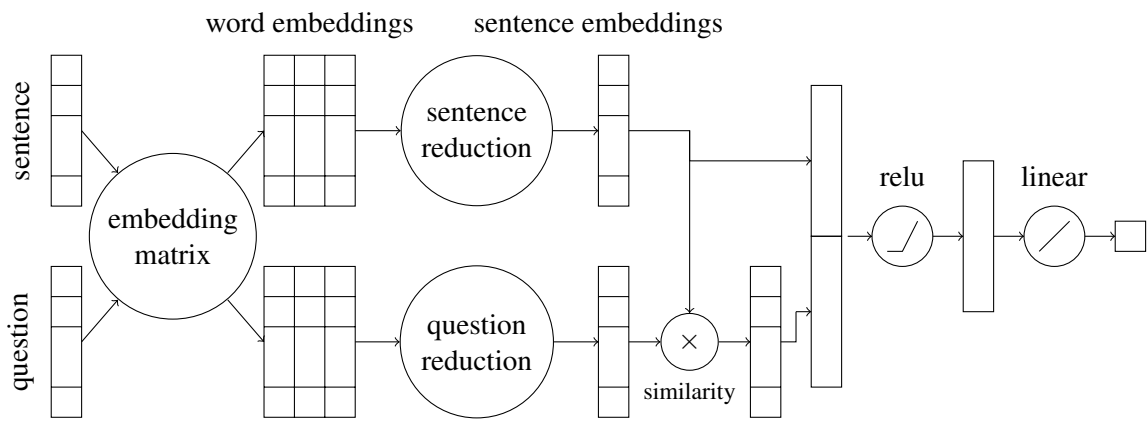

Figure~\ref{fig:nnr} shows the general architecture of the deep
learning systems explored in our experiments.
In a pre-processing stage, and not shown in the figure, the main text
of the source PubMed abstracts is split into sentences by using the
default NLTK\footnote{\url{http://www.nltk.org}} sentence
segmenter. The candidate sentences and questions undergo a simple
preprocessing stage that removes punctuation characters, and
lowercases the string and splits on blank spaces. Then, these are fed
to the system as a sequence of token identifiers. Figure~\ref{fig:nnr}
shows that the input to the system is a candidate sentence and the
question (as sequences of token IDs). The input is first converted to
sequences of word embeddings by applying an embedding matrix. The word
embedding stage is followed by a sentence and question reduction stage
that combines the word embeddings of each sentence into a sentence
embedding. Then, the sentence embedding and the question embedding are
compared by applying a similarity operation, and the vector resulting
from the comparison is concatenated to the sentence embedding for a
final regression comprising of a hidden layer of rectilinear
units (relu) and a final linear combination.

The weights of all stages are optimised by backpropagation in order to
minimise the MSE of the predicted score at
training time. Our experiments varied on the approach for sentence and
question reduction, and the approach to incorporate the similarity
between sentence and question, as described below.

To produce word embeddings we use word2vec, trained on a collection of
over 10 million PubMed abstracts as described in previous
sections. The resulting word embeddings are encoded in the embedding
matrix of Figure~\ref{fig:nnr}. We experimented with the
possibility of adjusting the weights of the embedding matrix by
backpropagation, but the results did not improve. The results reported
in this paper, therefore, used a constant embedding matrix. We
experimented with various sizes of word embeddings and chose 100 for
the experiments in this paper.

After obtaining the word embeddings, we experimented with the
following approaches to produce the sentence vectors:

\begin{description}
\item[Mean:] The word embeddings provided by word2vec map words into a
  dimensional space that roughly represents the word meanings, such
  that words that are similar in meaning are also near in the embedded
  space. This embedding space has the property that some semantic
  relations between words are also mapped in the embedded space
  \cite{Mikolov:2013}. It is therefore natural to apply vector
  arithmetics such as the sum or the mean of word embeddings of a
  sentence in order to obtain the sentence embedding. In fact, this
  approach has been used in a range of applications, on its own, or as
  a baseline against which to compare other more sophisticated
  approaches to obtain word embeddings, e.g. work by \newcite{Yu:2014}
  and \newcite{Kageback:2014}. To accommodate for different sentence
  lengths, in our experiments we use the mean of word embeddings
  instead of the sum.
\item[CNN:] Convolutional Neural Nets (CNN) were originally developed
  for image processing, for tasks where the important information may
  appear on arbitrary fragments of the image \cite{Fukushima:1980}. By
  applying a convolutional layer, the image is scanned for salient
  information. When the convolutional layer is followed by a maxpool
  layer, the most salient information is kept for further processing.

  We follow the usual approach for the application of CNN for word
  sequences, e.g. as described by \newcite{Kim:2014}. In particular, the
  embeddings of the words in a sentence (or question) are arranged in
  a matrix where each row represents a word embedding. Then, a set of
  convolutional filters are applied. Each convolutional filter uses a
  window of width the total number of columns (that is, the entire
  word embedding). Each convolutional filter has a fixed height,
  ranging from 2 to 4 rows in our experiments. These filters aim to
  capture salient ngrams. The convolutional filters are then followed
  by a maxpool layer.

  Our final sentence embedding concatenates the output of 32 different
  convolutional filters, each at filter heights 2, 3, and 4. The
  sentence embedding, therefore, has a size of $32\times3 = 96$.

\item[LSTM:] The third approach that we have used to obtain the
  sentence embeddings is recurrent networks, and in particular Long
  Short Term Memory (LSTM).  LSTM has been applied successfully to
  applications that process sequences of samples
  \cite{Hochreiter:1997}.  Our experiments use TensorFlow's
  implementation of LSTM cells as described by \newcite{Pham:2013}.

  In order to incorporate the context on the left and right of each
  word we have used the bidirectional variant that concatenates the
  output of a forward and a backward LSTM chain.  As is usual
  practice, all the LSTM cells in the forward chain share a set of
  weights, and all the LSTM cells in the backward chain share a
  different set of weights. This way the network can generalise to an
  arbitrary position of a word in the sentence. However, we expect
  that the words of the question behave differently from the words of
  the candidate sentence. He have therefore used four distinct sets of
  weights, two for the forward and backward chains of the candidate
  sentences, and two for the question sentences.

  In our experiments, the size of the output of a chain of LSTM cells
  is the same as the number of features in the input data, that is,
  the size of the word embeddings. Accounting for forward and backward
  chains, and given word embeddings of size 100, the size of the final
  sentence embedding is 200.
\end{description}

Figure~\ref{fig:nnr} shows how we incorporated the similarity between
the question and the candidate sentence. In particular, we calculated
a weighted dot product, where the weights $w_i$ can be learnt by
backpropagation:

$$sim(q,s) = \sum_iw_iq_is_i$$

Since the sum will be performed by the subsequent relu layer, our
comparison between the sentence and the question is implemented as a
simple element-wise product between the weights, sentence embeddings,
and question embeddings.

An alternative similarity metric that we have also tried is as
proposed by \newcite{Yu:2014}. Their similarity metric allows for
interactions between different components of the sentence vectors, by
applying a $d\times d$ weight matrix $W$, where $d$ is the sentence
embedding size, and adding a bias term:

$$simYu(q,s) = q^TWs+b$$

In both cases, the optimal weights and bias are learnt by
backpropagation as part of the complete neural network model of the
system.

Table~\ref{tab:nnr-results} shows the average MSE
of 10-fold cross-validation over the training data of BioASQ~5b.
\begin{table}
  \centering
% \begin{tabular}{lrrr}
% \emph{Method} & \emph{Plain} & \emph{Sim} & \emph{SimYu}\\
% \midrule
% Constant       & 0.00384\\
% Tf.idf         & 0.00349 & 0.00342\\
% \midrule
% Mean           & 0.00347 & 0.00338 & \textbf{0.00336}\\
% LSTM           & 0.00345 & 0.00341 & 0.00339\\
% CNN            & 0.00352 & 0.00347 & 0.00347\\
% \end{tabular}
\begin{tabular}{lrrr}
\emph{Method} & \emph{Plain} & \emph{Sim} & \emph{SimYu}\\
\midrule
%Constant       & 0.00384\\
Tf.idf         & 0.00354\\
\midrule
SVD            & 0.00345 & 0.00334 & 0.00342\\
\midrule
Mean           & 0.00341 & 0.00330 & 0.00331\\
CNN            & 0.00350 & 0.00348 & 0.00349\\
LSTM           & 0.00344 & 0.00335 & 0.00336\\
\end{tabular}
  \caption{Average MSE of 10-fold cross-validation.}
  \label{tab:nnr-results}
\end{table}
``Tf.idf'' is a neural network with a hidden layer of 50 relu cells,
followed by a linear cell, where the inputs are the \emph{tf.idf} of
the words. ``SVD'' computes the sentence vectors as described in
Section~\ref{sec:simple}, with the only difference being that now we
chose 100 SVD components (instead of 200) for comparison with the
other approaches shown in Table~\ref{tab:nnr-results}.

We observe that all experiments perform better than the Tf.idf
baseline, but there are no major differences between the use of SVD
and the three approaches based on word embeddings. The systems which
integrated a sentence similarity performed better than those not using
it, though the differences when using CNN are negligible. Each cell in
Table~\ref{tab:nnr-results} shows the best results after grid searches
varying the dropout rate and the number of epochs during training.

For the ``nnr'' run, we chose the combination ``Mean'' and ``Sim'' of
Table~\ref{tab:nnr-results}, since they produced the best results in
our experiments (although only marginally better than some of the
other approaches shown in the table).

\section{Submission Results}

At the time of writing, the human evaluations had not been released,
and only the ROUGE results of all 5 batches were
available. Table~\ref{tab:results} shows the F1 score of ROUGE-SU4.
\begin{table*}
  \centering
  \begin{tabular}{llllll}
   \textbf{System} & \textbf{Batch 1}& \textbf{Batch 2} &
                                                          \textbf{Batch
                                                          3} &
                                                               \textbf{Batch
                                                               4} &
                                                                    \textbf{Batch 5}\\
\midrule
\textbf{trivial} & \textbf{0.5498} & 0.4901 & 0.5832 & 0.5431 & 0.4950\\
\textbf{simple} & 0.5068 & \textbf{0.5182} &  \textbf{0.6186} & \textbf{0.5769} & \textbf{0.5840}\\
\textbf{regression} & 0.5186 & 0.4795& 0.5785 & 0.5436 & 0.4784\\
\textbf{nnr} & 0.4192 & 0.3920 & 0.5196 & 0.4445 & 0.4000\\
  \end{tabular}
  \caption{ROUGE-SU4 of the 5 batches of BioASQ 2017.}
  \label{tab:results}
\end{table*}

\begin{figure}
  \centering
  \includegraphics[width=\columnwidth]{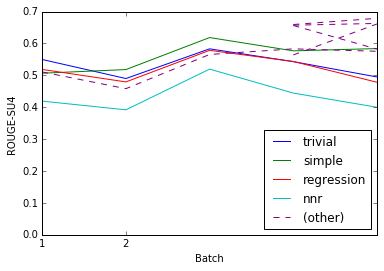}
  \caption{Top ROUGE-SU4 scores of the 5 batches of BioASQ 2017.}
  \label{fig:results}
\end{figure}

Figure~\ref{fig:results} shows the same information as a plot that
includes our runs and all runs of other participating
systems with higher ROUGE scores. The figure shows that, in the first
three batches, only one run by another participant was among our
results (shown as a dashed line in the figure). Batches 4 and 5 show
consistent results by our runs, and improved results of runs of other
entrants.

The results are consistent with our experiments, though the absolute
values are higher than those in our experiments. This is probably
because we used the entire training set of BioASQ 5b for our
cross-validation results, and this data is the aggregation of the
training sets of the BioASQ tasks of previous years. It is possible
that the data of latter years are of higher quality, and it might be
useful to devise learning approaches that would account for
this possibility.

\section{Conclusions}

At the time of writing, only the ROUGE scores of BioASQ 5b were
available. The conclusions presented here, therefore, do not
incorporate any insights of the human judgements that are also part of
the final evaluation of BioASQ.

Our experiments show that a trivial baseline system that returned the
first $n$ snippets appears to be hard to beat. This implies that the
order of the snippets matters. Even though the judges were not given
specific instructions about the order of the snippets, it would be
interesting to study what criteria they used to present the snippets.

Our runs using regression were not significantly better than simpler
approaches, and the runs using deep learning reported the lowest
results. Note, however, that the input features used in the runs using
deep learning did not incorporate information about the
snippets. Table~\ref{tab:nnr-results} shows that the results using
deep learning are comparable to results using tf.idf and using SVD, so
it is possible that an extension of the system that incorporates
information from the snippets would equal or better the other systems.

Note that none of the experiments described in this paper used
information specific to the biomedical domain and therefore the
methods described here could be applied to any other domain.

\section*{Acknowledgments}

Some of the experiments in this research were carried out in cloud
machines under a Microsoft Azure for Research Award.

\bibliography{bioasq2017}
\bibliographystyle{acl_natbib}

\end{document}